\documentclass[final,5p,times,twocolumn,numbers,sort&compress]{elsarticle}

\usepackage{graphicx}
\usepackage{amssymb}
\usepackage{amsmath,amsfonts,amsthm}
\usepackage{mathrsfs}
\usepackage{multirow}
\usepackage{booktabs}
\usepackage{algorithm}
\usepackage{algorithmicx}
\usepackage{algpseudocode}
\usepackage{listings}
\usepackage{csquotes}
\usepackage{xcolor}
\usepackage{textcomp}

\usepackage[version=4]{mhchem}
\usepackage{siunitx}
\usepackage{array}
\usepackage{verbatim}
\usepackage[hyphens]{url}
\usepackage{hyperref}
\hypersetup{breaklinks=true}
\expandafter\def\expandafter\UrlBreaks\expandafter{\UrlBreaks\do\-\do\_\do\a\do\b\do\c\do\d\do\e\do\f\do\g\do\h\do\i\do\j\do\k\do\l\do\m\do\n\do\o\do\p\do\q\do\r\do\s\do\t\do\u\do\v\do\w\do\x\do\y\do\z\do\A\do\B\do\C\do\D\do\E\do\F\do\G\do\H\do\I\do\J\do\K\do\L\do\M\do\N\do\O\do\P\do\Q\do\R\do\S\do\T\do\U\do\V\do\W\do\X\do\Y\do\Z\do0\do1\do2\do3\do4\do5\do6\do7\do8\do9}
\usepackage{subcaption}
\usepackage[normalem]{ulem}

\theoremstyle{plain}

\theoremstyle{definition}

\journal{Applied Energy}

\begin{document}

\begin{frontmatter}

%% Title, authors and addresses
\title{Reducing electric grid emissions using flexible data centers: a price-aware cherry-picking approach for effective load shaping}

\author[google]{Bokan Chen\fnref{equal}}
\author[google]{Raiden Hasegawa\fnref{equal}}
\author[google]{Adriaan Hilbers\fnref{equal}}
\author[google]{Ross Koningstein\fnref{equal}}
\author[google]{Ana Radovanovi\'{c}\corref{cor1}\fnref{equal}}
\ead{anaradovanovic@google.com}
\author[google]{Utkarsh Shah\fnref{equal}}
\author[google]{Gabriela Volpato\fnref{equal}}
\author[epe]{Mohamed Ahmed\fnref{equal}}
\author[epe]{Tim Cary\fnref{equal}}
\author[epe]{Rod Frowd\fnref{equal}}

\cortext[cor1]{Corresponding author}
\fntext[equal]{The author names are ordered alphabetically, first Google affiliated and then Electric Power Engineers affiliated.}

\affiliation[google]{organization={Google},
            country={USA}}
\affiliation[epe]{organization={Electric Power Engineers},
            country={USA}}

\begin{abstract}
Shaping multi-megawatt loads, such as data centers, impacts generator dispatch on the electric grid, which in turn affects system \ce{CO2} emissions and energy cost. Substantiating the effectiveness of prevalent load shaping approaches, such as those based on grid-level average carbon intensity, locational marginal price (LMP), or marginal emissions, is challenging due to the lack of real-world grid counterfactuals required for accurate attribution. Because full-scale market clearing models and proprietary generator bid data are rarely accessible outside of grid operators, prior studies have largely relied on historical signal correlations without dispatch counterfactuals, geographically coarse zonal models, or small synthetic test networks. To address this gap, this study uses a close replica of ERCOT's day-ahead Security Constrained Unit Commitment (SCUC) and Security Constrained Economic Dispatch (SCED) market clearing process (spanning nearly 8,900 buses) for counterfactual analysis of the impact of a broad set of load shaping approaches on grid \ce{CO2} emissions and cost of electricity. In terms of annual grid-level \ce{CO2} emissions reductions, LMP-based shaping outperforms other common approaches, but can be significantly improved upon. Examining the performance of practicable approaches, which rely only on observable or predictable day-ahead signals, motivates a more effective load shaping strategy: one that ``cherry-picks'' a daily approach based on observable grid signals and historical data. The overall yearly grid-level \ce{CO2} emissions savings of cherry-picking can be up to 2--3 times larger than LMP-based shaping, without increasing the cost when compared to no load shaping. Motivated by shifting compute load between data centers, we also simulate the temporal and spatial shifting of energy between grid nodes and show that shifting consumption to ``greener'' and ``cheaper'' grid zones more effectively reduces grid-level \ce{CO2} emissions by leveraging higher renewable energy production of ``greener'' grid zones and reducing load in mainly fossil fuel supplied zones. We show that \ce{CO2} emissions savings from load shifting between two grid nodes can be 70\% larger than shifting energy consumption only in time at each location separately. In addition to data centers, the proposed load shaping strategy is applicable to any large flexible consumer capable of changing its power consumption as discussed in this article, and the principled approach to design of the cherry-picking rule can be extended to any grid. Finally, because load shaping based on regional average carbon intensity can increase grid-level $\ce{CO2}$ emissions, our findings highlight the need for Scope 2 standards and corporate accounting frameworks to adopt locational, price-sensitive, and congestion-aware signals.
\end{abstract}

\begin{keyword}
Power grid \sep Data center \sep Load shifting \sep \ce{CO2} emissions reduction
\end{keyword}

\end{frontmatter}

%% main text

\section{Introduction}\label{sec1}

The United States is currently experiencing a marked acceleration in electricity demand. Studies show aggregated US winter peak load could grow 21.5\% over the next decade \citep{NERC2024LTRA}. Much of this growth can be attributed to the rapid growth of data center demand. According to Shehabi et al.~\citep{Shehabi2024DataCenter}, data centers consumed 4.4\% of total U.S. electricity in 2023 and are expected to consume approximately 6.7\% to 12\% by 2028.

This growth highlights the importance of load flexibility for demand peak shaving (saving infrastructure costs), grid reliability and decarbonization. Load flexibility alleviates some immediate pressure on generation capacity and transmission infrastructure \citep{norris2025rethinking}. By modulating demand, flexible loads can make use of more renewable energy, and through demand reduction services (e.g. demand response) improve grid stability and resiliency \citep{Murphy2024DemandFlexibility}. In California, load shifting is reported to mitigate challenges like the \enquote{duck curve}, which reduces ramping of thermal generators, curtailment of renewable generation and peak capacity needs \citep{Gallo2018MOBILIZINGTHE}. 

\subsection{Carbon-aware load shaping} \label{sec:literature}

While large compute providers have traditionally pursued their climate-related targets through clean energy investments in Power Purchase Agreements (PPAs) \citep{IEA2023DataCentres}, the role of data center flexibility in reducing grid-level emissions and achieving company-level carbon goals has received growing attention. Demand flexibility, especially shaping loads at the multi-megawatt level, affects electricity generation dispatch on the grid and the associated \ce{CO2} emissions and energy cost. Thus, load flexibility of data centers, and more broadly large fleets of flexible load aggregations referred to as Virtual Power Plants (VPPs), could also be used to reduce grid-level \ce{CO2} emissions. In practice, carbon-aware load shaping is typically performed by adapting electricity consumption in response to a signal that characterizes the carbon intensity of electricity across different times and locations, shifting flexible demand toward hours or nodes with lower reported carbon intensity \citep{radovanovic2022carbon, lindberg2021guide, CarbonAwareComputing2023, sukprasert2024implications}. However, because there is no single well-defined and universally agreed-upon carbon intensity signal (and such a universal signal likely does not exist), operators choose the guiding signal to follow. 

While flexible loads have traditionally participated in price-based demand response programs to reduce electricity costs and support grid reliability \citep{7360934, Mehra2023DemandResponse, Terrell2025FlexibleDataCenters}, major technology companies---including Google \citep{radovanovic2022carbon} and Microsoft \citep{Hauglie2023Xbox, CarbonAwareComputing2023}---have also deployed carbon-aware load shifting to advance their sustainability goals, typically relying on open or commercially available carbon data providers, such as Electricity Maps \citep{ElectricityMaps} and WattTime \citep{watttime2021watttime}, to reduce their reported Scope 2 carbon footprint \citep{GHGProtocol2015Scope2}. More generally, power load shaping in nodes of the electricity transmission grid based on signals such as average grid carbon intensity, locational marginal price (LMP), marginal carbon intensity, grid and local (zonal) wind and solar generation, etc., have been proposed and implemented, with hoped-for impact \citep{lindberg2021guide, sukprasert2024implications}. From historical grid telemetry alone, the effectiveness and limitations of such signals cannot be evaluated, as telemetry only records the dispatch outcomes that actually occurred, while counterfactual simulations on small synthetic test networks \citep{lindberg2021guide, gorka2024electricityemissions} do not capture the scale and market structure of actual regional grids. Evaluating the true system-wide \ce{CO2} emissions impact of shifting a multi-megawatt load requires modeling how market-clearing optimization dispatches generation across the transmission network with and without the shift. This impact cannot be captured simply by multiplying a shifted load profile by historical or forecasted carbon intensity signals.

A substantial body of literature has examined how carbon intensity signals and flexible demand interact with power system operations and affect long-run decarbonization. Foundational empirical studies quantified short-run marginal emissions factors across U.S.\ regions and time periods using historical generation data \citep{siler2012marginal}, and developed marginal operating emissions rate models to evaluate how displaced emissions vary across locations and renewable or demand-side resources \citep{callaway2018location}, together forming the methodological basis for commercial marginal carbon signals such as WattTime \citep{watttime2021watttime}. Recent studies have also investigated long-run marginal emission rates that account for induced structural changes in grid capacity \citep{gagnon2022planning}, as well as the system-level impacts of flexible loads and hourly carbon-free energy matching on achieving net-zero power systems \citep{ricks2023minimizing, xu2024system}. In the context of data centers specifically, in \cite{lindberg2021guide}, the authors proposed a geo-shifting model for data centers and studied its impact on grid carbon emissions by using different carbon costs. In \cite{riepin2025spatio}, the authors proposed a way for large corporations to connect their PPA investments and their ability to shift data center load both spatially and temporally, to achieve \enquote{24/7 carbon free energy} (CFE) matching \citep{IEA2023DataCentres}. 

Carbon accounting and \ce{CO2} emissions impact represent fundamentally different perspectives—one is an administrative reporting framework, and the other is a physical measurement of real-world grid operations. One of the most common carbon accounting approaches calculates carbon footprint using the grid-level {\it average carbon intensity} \citep{GHGProtocol2015Scope2} and is referred to as Location-Based Accounting. For example, Google \citep{radovanovic2022carbon} uses this signal and performs load shaping to minimize its data centers' carbon footprint. While average carbon intensity serves carbon accounting well by capturing aggregate generator emissions, it is suboptimal for day-ahead load shifting: it ignores physical grid constraints \citep{gorka2024electricityemissions} and most notably does not anticipate how the grid responds to large-scale load changes. Alternatively, {\it locational marginal emissions} is a metric designed to quantify the change in overall system emissions that results from a small change in electricity demand at a specific location and time. When used as a day-ahead load shaping signal, studies reveal that it effectively reduces grid carbon emissions with small load shifts \citep{gorka2024electricityemissions}. However, the calculation of this signal depends on detailed information about the grid that is often not publicly available \citep{lindberg2021guide}. Where it is available, e.g.\ in PJM \citep{PJM2024MarginalEmissions}, it can be highly volatile between time steps and prone to changes with large load shifts \citep{gorka2024electricityemissions}, making it inappropriate for load shifting. Researchers also explored a metric called {\it software carbon intensity}, which combines the locational marginal carbon emissions of the grid, the energy consumed, and the scope 3 carbon emissions of the hardware device running the software \citep{dodge2022measuring, CarbonAwareComputing2023}.

Importantly, different carbon accounting metrics are not fully aligned with each other. Optimizing one metric can lead to the deterioration of another \citep{gorka2024electricityemissions, sukprasert2024implications}, and optimizing the carbon footprint of individual flexible energy assets does not automatically lead to reduction in grid-level \ce{CO2} emissions. In fact, there is a critical distinction between reducing the carbon footprint of an individual consumer (i.e., the flexible load), reducing operational emissions of the overall grid, and achieving long-term structural decarbonization. When and where electricity is consumed impacts generation dispatch. While flexible demand can reduce emissions in the short term, it does not automatically result in the decommissioning of high-emission generation assets or a transition toward cleaner generation portfolios. Indeed, while short-term operational studies find that locational marginal emission signals reduce dispatch emissions more effectively than average carbon intensity \citep{lindberg2021guide, gorka2024electricityemissions}, long-term capacity-expansion modeling indicates that short-run marginal emission rates can perform worse than average or long-run marginal emission rates over multi-year horizons because they treat the generation fleet as fixed and neglect induced renewable capacity additions \citep{gagnon2022planning}. Without aligned market structures and policy frameworks that incentivize investment in low-carbon dispatchable generation and infrastructure and the retirement of legacy fossil fuel plants, grids can remain vulnerable to both fragility and persistent emissions intensity \citep{angwin2020shorting}. 

\subsection{Scope of study}

In this article, we study the impact of a set of approaches to day-ahead temporal and spatial load shifting on grid-level \ce{CO2} emissions. To our knowledge, this is the first study to systematically evaluate load-shaping approaches across diverse grid signals using counterfactuals derived from granular ERCOT market data, where day-ahead Security Constrained Unit Commitment (SCUC) and Security Constrained Economic Dispatch (SCED) models are calibrated against the recorded 2023 dispatch.

Prior studies have investigated carbon-aware load shifting using empirical comparisons of region-wide carbon intensity signals \citep{sukprasert2024implications}, coarse zonal capacity-expansion models \citep{riepin2025spatio}, or dispatch counterfactuals on synthetic test networks with stylized market-clearing models \citep{lindberg2021guide, gorka2024electricityemissions}. While counterfactual studies on synthetic networks established that the system-wide impacts of carbon-aware load shifting are non-trivial and that following certain signals (such as regional average carbon intensity or excess low-carbon capacity) can sometimes backfire by increasing system-wide carbon emissions \citep{lindberg2021guide, gorka2024electricityemissions}, such stylized models cannot directly guide practical load-shifting operations. Designing an operational strategy requires anticipating how a specific regional grid responds to large-scale load changes using realistic network and market data, while accounting for the practical availability of day-ahead signals, which constrains algorithm design. To bridge this gap, this study evaluates the \ce{CO2} and cost impact of commonly proposed day-ahead load shaping approaches and addresses how to select among available guiding signals using a close replica of ERCOT's day-ahead SCUC and SCED market clearing process \citep{luo2010security, EPEConsulting}. 

The load shaping is conducted at two $400$MW ERCOT transmission nodes in different weather zones with distinct generation mixes: NORTH and EAST. We assume that the two loads can change their power demand between $320$ and $480$MW on an hourly basis, while preserving total daily energy consumed. In addition, the simulator was also modified to run a causal optimization-based approach with the explicitly incorporated, scaled generators' \ce{CO2} emissions term in the objective function (see \ref{sbsec:simulation}). This version of the market-clearing solver was used to establish an \textit{optimization-based benchmark}—identifying daily load shapes that provide an upper bound on \ce{CO2} reductions for impact ranking. However, this benchmark is unachievable in practice because an individual load operator lacks access to the system operator's proprietary network model, contingency constraints, and confidential generator bid curves.

\subsection{Contributions}

By combining this systematic evaluation framework with a granular and calibrated model of ERCOT's day-ahead SCUC and SCED market-clearing process, this work delivers the following key novel contributions:

\begin{itemize}
\item We show that no single approach reaches the {\it optimization-based benchmark} in terms of average daily \ce{CO2} emissions savings. Furthermore, the causal simulation framework across different load shaping approaches reveals that there is no implementable single signal that is the most effective every day. Conditions under which implementable approaches do well or particularly poorly provide clues for larger impact.

\item A new adaptive, practicable daily \enquote{cherry-picking} approach is proposed and demonstrated to outperform any single-signal load shaping in terms of \ce{CO2} emissions savings. The most effective features for the next day's signal selection are determined to be local grid signals of the flexible load, such as the locational marginal price (LMP), zonal renewable generation and the zone itself, with its implicit characteristics, such as the generation mix, topology, and characteristic bidding of its generators.

\item Empirical validation using a high-fidelity market simulator calibrated to historical 2023 ERCOT day-ahead SCUC and SCED clearing data demonstrates that the proposed \enquote{cherry-pick} strategy achieves 2 to 3 times the annual \ce{CO2} savings of the best single-signal approach, while inducing negligible wholesale market price impacts ($\le$ 0.6\$/MWh). Importantly, we demonstrate, in Subsection \ref{sbsec:single_node_eval}, that the approach selection rule is not \textit{too} sensitive to the local grid signal values, which ensures the implementability and effectiveness of cherry-picking in practice. Furthermore, this analysis highlights the importance of the location selection for \ce{CO2}-effective and cost-effective load shaping. Note that the previous insights hold irrespective of the type of the flexible load (e.g. data center, fleet of electric vehicles, EVs), and are applicable to different distributed energy resource (DER) use cases \citep{haberle2021control, haberle2023grid, marinescu2022dynamic, bjork2022dynamic}. 

\item Temporal and spatial shifting of computing between data centers achieves at least 2 times greater \ce{CO2} emissions reductions than single-node temporal shifting alone. Layering the proposed \enquote{cherry-pick} selection policy across the two selected nodes delivers more than 70\% increase in emissions savings (Subsection \ref{sec:2nodeshifting}).

\end{itemize}

\subsection{Outline}

The remainder of this paper is organized as follows. Section \ref{sec:loadshaping} describes the simulation setup used to study the impact of load shaping approaches based on a range of available grid signals. Section \ref{sec:cherrypicking_design} contains the analysis of the counterfactual dispatch outcomes across individual signals, identifies the key grid factors governing daily signal effectiveness, and derives a novel rule-based cherry-picking strategy, used to select the next day's load shaping approach. The details of the numerical evaluation of the simulation-tuned and data-driven cherry-picking rules can be found in Subsections \ref{sbsec:single_node_eval} and \ref{sec:datadriven}, respectively. The impact of price-driven load shaping approaches on \ce{CO2} emissions and generation dispatch can be found in Subsection \ref{sbsec:evimpact}, while the impact of multi-node spatial and temporal load shifting is examined in Subsection \ref{sec:2nodeshifting}. Section \ref{sec:conclusions} concludes the paper.

%%%%%%%%%%%%%%%%%%%%%%%%%
%%%%%%%%%%%%%%%%%%%%%%%%%
\section{Load shaping simulation setup} \label{sec:loadshaping}

In this section, we describe the simulation setup used to evaluate day-ahead carbon-aware load shaping in ERCOT, detailing the flexible data center load model, the two transmission bus locations, the baseline and guiding signals, and the calibrated SCED simulator.

\begin{figure}[h]
  \centering
  \includegraphics[width=1\linewidth]{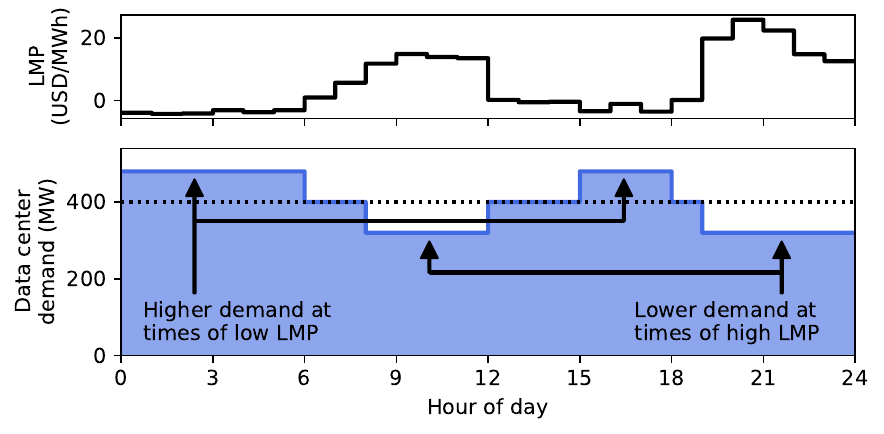}
  \caption{How a signal (in this case hourly LMP) translates into data center demand changes.}
  \label{fig:policy}
\end{figure}

{\bf Load flexibility modeling}: To assess the impact of carbon-aware load shaping on overall grid emissions and electricity costs, we explore several different approaches to increase or decrease the flexible power load in the selected grid nodes. The one compatible with data center load modulation and most effective in reducing \ce{CO2} emissions uses max-nominal-min load changes. More specifically, for each hour $t \in \{1,\dots,24\}$ of the day, we assume that a data center may consume hourly power $P_t \in \{320, 400, 480\}$\,MW, with the fixed total daily energy equal to \(400\times 24 = \SI{9600}{\mega\watt h}\). Furthermore, power shaping is such that the load consumes 320MW and 480MW for 9 hours each, and 400MW for 6 hours (``up'', ``down'' or unchanged), satisfying
\begin{equation}
\begin{aligned}
\sum_{t=1}^{24} P_t &= \SI{9600}{\mega\watt h}, \quad P_t \in \{320, 400, 480\}\,\text{MW}, \\
\bigl|\{t : P_t = 480\}\bigr| &= \bigl|\{t : P_t = 320\}\bigr| = 9, \quad \bigl|\{t : P_t = 400\}\bigr| = 6.
\end{aligned} \label{eq:load_shaping}
\end{equation}
Given a day-ahead guiding signal $s = (s_1,\dots,s_{24})$, each single-signal shaping policy ranks the 24 hours by $s_t$ and assigns $P_t = 480$\,MW to the 9 most favorable hours (lowest $s_t$ for cost or carbon-intensity signals; highest $s_t$ for renewable-generation signals), $P_t = 320$\,MW to the 9 least favorable hours, and $P_t = 400$\,MW to the remaining 6 hours. This flexibility model, while simplistic, is achievable in large data centers: in particular, it is directly enabled by the day-ahead capacity-curve mechanism for flexible workload admission and demand response that we developed and deployed in production across Google's fleet of data centers \citep{radovanovic2022carbon, Mehra2023DemandResponse, Terrell2025FlexibleDataCenters}, or potentially some other mechanism. Each analyzed load shaping approach in this paper uses this load changing variant. Figure \ref{fig:policy} shows one such example.

\begin{figure}[h]
  \centering
  \includegraphics[width=1\linewidth]{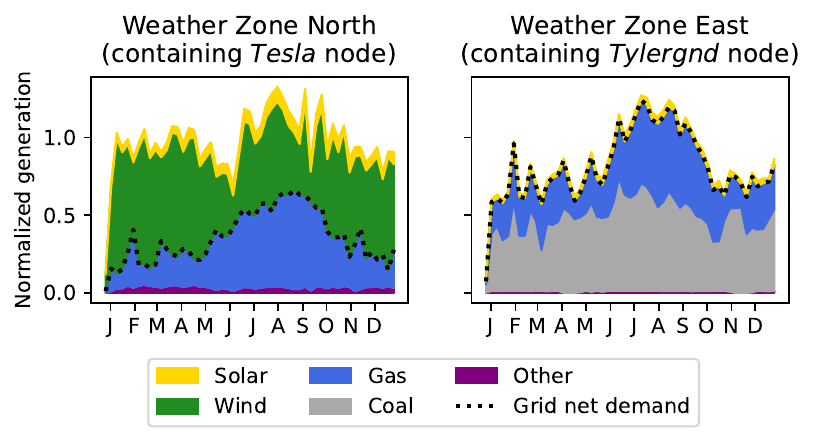}
  \caption{Generation mix in North and East ERCOT weather zones in 2023. The dominant types of dispatched generation are wind and solar, st (steam turbine) coal, combined cycle and non-combined cycle gas generations, i.e. st (steam turbine) gas, ct (combustion turbine) gas, and ic (internal combustion) gas. Other sources of energy include nuclear, hydro, battery storage, biomass.}
  \label{fig:zgen}
\end{figure}

{\bf Load shaping locations}: The impact of load shaping is evaluated in scenarios with one and two flexible loads in ERCOT grid nodes (i.e. buses). One of the two flexible loads is in the NORTH WEATHER ZONE, at the bus named Tesla. This is the ERCOT zone with the largest renewable generation supply, and is a location of Google's data center. The second location was selected in the EAST WEATHER ZONE, at the bus named Tylergnd. This zone has plentiful coal and gas, and minimal renewable production (see Figure \ref{fig:zgen}).

While the main focus of the study is to evaluate the impact of temporal shifting using the commonly discussed load shaping approaches, the impact of spatial and temporal shifting of energy consumption between the two considered flexible loads is also included to address recent developments in data center compute shifting \citep{fridgen2017shifting,radovanovic2022carbon}.

{\bf Baseline load benchmark}: The benchmark for comparison of policies is a flat load profile (referred to as baseline) of \SI{400}{\mega\watt}, typically used to evaluate the impact of load shaping \citep{riepin2025spatio, radovanovic2022carbon}.

{\bf Load shaping signals}: It is assumed that a flexible load operator knows (or has reliable day-ahead predictions of) grid signals used for load shaping: (i) \textit{avg}, grid average carbon intensity, (ii) \textit{base}: ``flat'' load regime as the baseline for evaluating the impact, (iii) \textit{cfe}, purchased carbon-free energy, (iv) \textit{overnight} (labeled \textit{nev} in Figure \ref{fig:total-impact-over-year}), increases load at night, and reduces it during the high demand hours within a day, (v) \textit{lme}, locational marginal emissions, (vi) \textit{lmp}, locational marginal price (LMP), (vii) \textit{ws}, grid's intermittent renewable generation, wind and solar, (viii) \textit{wme}, WattTime \citep{watttime2021watttime}, (ix) \textit{zws}, zonal intermittent renewable generation, wind and solar. Locational marginal emissions (\textit{lme}) in this paper are computed numerically using the simulation results; because \textit{lme} depends on the details of the market settlement process and its outcomes, which are not publicly available, it is not implementable in practice.

{\bf ERCOT SCED simulation model}: EPE's granular and calibrated direct current optimal power flow (DC-OPF) simulations, which emulate ERCOT's day-ahead SCED generation dispatch and market settlements, are used to compute the optimal dispatch using the shaped daily data center power demand as the input (see \ref{sec-appendix} for more details). In addition, the optimization's objective is modified to include \ce{CO2} emissions of dispatched generation, and used to compute the emissions-optimal daily load shapes; as discussed in \ref{sbsec:simulation}, this optimization-based approach relies on the full, granular DC-OPF optimization setup proprietary to EPE (and is therefore also not implementable in practice), and is used to establish the upper-bound benchmark against which the impact of different approaches is compared. The \ce{CO2} emissions are computed using the following estimates for carbon intensity across generation types (based on data from Electricity Maps \citep{ElectricityMaps}, all in $\text{gCO}_\text{2}$/$\text{kWh}$): $11$ (Wind), $45$ (Solar), $12$ (Nuclear), $820$ (Coal), $490$ (all Gas technologies), $230$ (Biomass), $24$ (Hydro).

%%%%%%%%%%%%%%%%%%
%%%%%%%%%%%%%%%%%%
\section{Cherry-picking for carbon-aware load shaping} \label{sec:cherrypicking_design}

In this section, we develop a practical day-ahead load shaping strategy focused on reducing grid-scale \ce{CO2} emissions by attributing emissions changes to shaped load changes. To do so, we first run counterfactual simulations across individual signals (Subsection \ref{sec:comparison}), next identify the key grid factors governing daily signal effectiveness (Subsection \ref{sec:features_alt}), and finally formulate our rule-based cherry-picking strategy (Subsection \ref{sec:policy}).

\subsection{Comparative evaluation of load shaping signals} \label{sec:comparison}\label{sbsec:yearperf}

Load shaping, particularly of larger loads, has an effect on dispatch of the various generators on the grid, and this in turn impacts both \ce{CO2} emissions and energy cost. As described above, a series of ERCOT DC-OPF simulation experiments are used to evaluate the impact of shaping. Specifically, across the full year 2023, we compare how each load shaping signal impacts annual grid-level \ce{CO2} emissions and wholesale electricity costs at both the renewable-rich (Tesla) and fossil-fuel-dominated (Tylergnd) locations.

\begin{figure}[h]
  \centering
  \includegraphics[width=\linewidth]{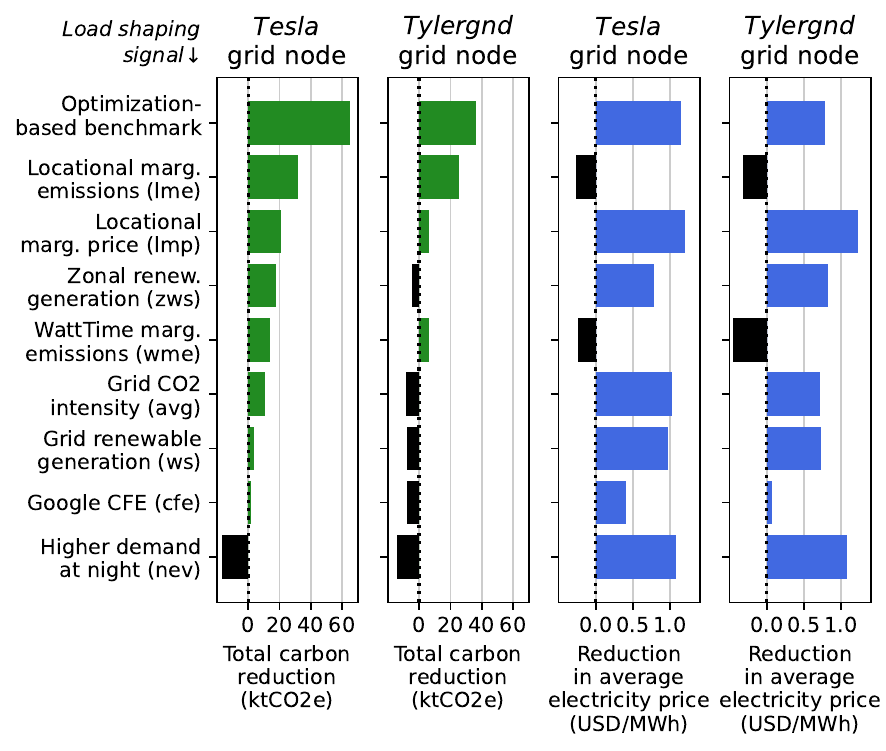}
  \caption{Reduced annual grid-level carbon emissions and average electricity price from different load shaping approaches.}
\label{fig:total-impact-over-year}
\end{figure}

For each daily load shape, the corresponding DC-OPF generation dispatch was used together with the carbon intensity of different generation types to compute the grid-level \ce{CO2} emissions compared to the baseline, \enquote{flat load} regime. Similarly, we evaluate the impact of load shaping on average day-ahead energy cost per unit of energy demand.

The simulation results in case of Tesla's demand shaping demonstrate that the largest yearly \ce{CO2} emissions savings among the considered implementable load shaping approaches have \textit{lmp}, \textit{zws}, and \textit{wme}, where they also reduce average grid energy cost rate. On the other hand, in the case of Tylergnd's demand shaping, no load shaping approach significantly outperforms the other ones with regards to the annual emissions savings, while their daily savings and average grid costs can be vastly different. For more details, see the yearly performance summary in Figure \ref{fig:total-impact-over-year}.

As expected, the non-implementable, {\it optimization-based} load shapes perform better than any commonly considered load shaping approach. Furthermore, consistent with Gorka et al.~\citep{gorka2024electricityemissions} (see Figure \ref{fig:total-impact-over-year}), \textit{lme} (also non-implementable in practice) demonstrates larger yearly \ce{CO2} emissions savings than other practically feasible load shapes at a price of increasing the average grid's wholesale energy cost rate due to a few extremely high-cost days. A commonly proposed approach that uses the next day's average carbon intensity as a shaping signal, \textit{avg}, reduces 48\% less \ce{CO2} emissions per year than the \textit{lmp} approach, when the Tesla load is shaped. Moreover, using \textit{avg} for shaping the Tylergnd load increases \ce{CO2} emissions on an annual basis (see Figure \ref{fig:total-impact-over-year}).

The {\it location} of the flexible load significantly affects the load shaping impact. Based on the {\it optimization-based} benchmark, shaping demand in Tesla can reduce $\sim$80\% more \ce{CO2} emissions per year than shaping the load in Tylergnd (65.4kt\ce{CO2}/yr vs 36.4 kt\ce{CO2}/yr), and represents $\sim$0.04\% of total ERCOT grid \ce{CO2} emissions. Furthermore, the \ce{CO2} emissions savings upper bound from shaping the Tesla load equals $\sim$5.8\% of its location-based carbon footprint (computed as the product between average energy demand in Tesla and grid-level average carbon intensity for each hour, and then summed over all hours in 2023), and $\sim$3.2\% in case of Tylergnd shaping.

Load shaping also influences electricity prices. Viewed at the grid average level (defined as average LMP, weighted by node demand), most approaches reduce average electricity price, with \textit{lmp} and \textit{overnight} reducing prices the most. This is because they reduce demand when generation margin (available supply - demand) is low, and at these times prices are typically high. By comparison, marginal emissions based policies may increase demand at such times, increasing electricity prices.

In terms of the electricity cost paid by the data center itself (consumption times LMP at its grid node), the impact is more pronounced. For example, at the Tesla bus, the data center's total energy cost is reduced by 13\% (\textit{lmp}), 11\% (\textit{avg}) but increased by 6\% when following the \textit{lme} policy.

Finally, load shaping affects both generation levels and revenue earned by generators (sum of generation times LMP at their respective grid nodes). Normalized against the average electricity price, the \textit{lmp} approach \textit{decreases} coal and gas generator revenue (-0.7\%) but \textit{increases} it for wind (+0.8\%) and solar (+1.2\%). For \textit{avg}, we see similar changes: -0.6\% for coal, -0.4\% for gas, +0.6\% for wind and +1.8\% for solar. These changes come from a combination of reducing renewable curtailments (increasing their generation levels) and increasing prices at times of high renewable generation levels. Hence, besides the operational scale (affecting generation dispatch), such policies encourage decarbonization on the planning scale by improving the investment case for renewables over fossil fuel technologies and hence encouraging their buildout.

%%%%%%%%%%%%%%
%%%%%%%%%%%%%%
\subsection{Key factors influencing load shaping effectiveness} \label{sec:features_alt}
\begin{figure}[h]
  \centering
  \includegraphics[width=0.95\linewidth]{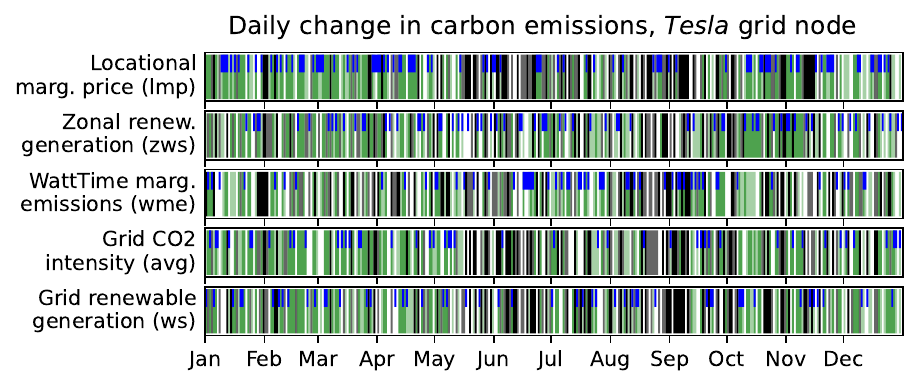}
  \includegraphics[width=0.95\linewidth]{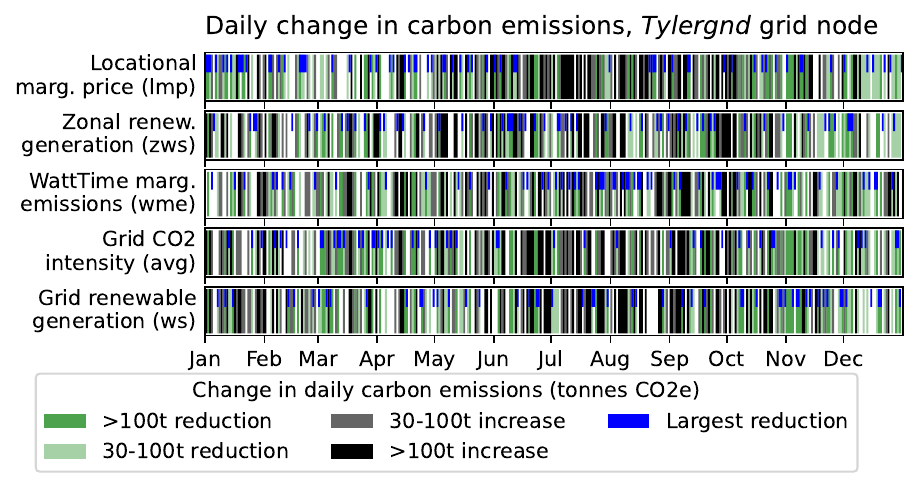}
  \caption{Daily grid-level \ce{CO2} emission reductions from the most effective approaches. \textit{lmp} reduces \ce{CO2} emissions on most days within the year. On the other hand, \textit{wme} is more effective during summer months from May to September.}
  \label{fig:dailyImpactAlt}
\end{figure}

Counterfactual simulation results presented in Figure \ref{fig:dailyImpactAlt} show variation in the effectiveness of different load shaping approaches on daily \ce{CO2} emissions. It is evident that no single load shaping approach consistently reduces \ce{CO2} emissions or reduces the energy cost rate more than any other approach each and every day. While the locational marginal price \textit{lmp} approach reduces both emissions and electricity costs on more days than any other single approach, its efficacy is highly variable, and there are many days when other approaches are more effective. To optimize impact, we analyze the connection between grid conditions and variables that govern the effectiveness of different load-shaping approaches on dispatched generators and the corresponding \ce{CO2} emissions.

This visual evidence in Figure \ref{fig:dailyImpactAlt} suggests an underlying seasonal correspondence with cost-effective generation dispatch. Seasons are reflected in the operation of the electric grid in several ways. The total grid demand and renewable capacity vary by season, and generation dispatch depends on the total grid demand and renewable capacity. 

\subsubsection{Grid net demand} \label{sbsec:gnd}

To quantitatively capture seasonal demand patterns, we calculated the Grid Net Demand (GND) by subtracting ERCOT-wide total (wind+solar) renewable generation from total grid demand. GND isolates the grid's reliance on non-renewables like gas, coal, nuclear, etc. In ERCOT, high GND is mainly driven by high demand, while low GND typically occurs outside the summer months when the total demand is generally lower. Power demand in the warmer months can be twice that during winter months. Wind and solar production in ERCOT fluctuates day to day, and is roughly 10\% lower in summer, but is relatively consistent throughout the year.

\begin{figure}[h]
  \centering
  \includegraphics[width=0.95\linewidth]{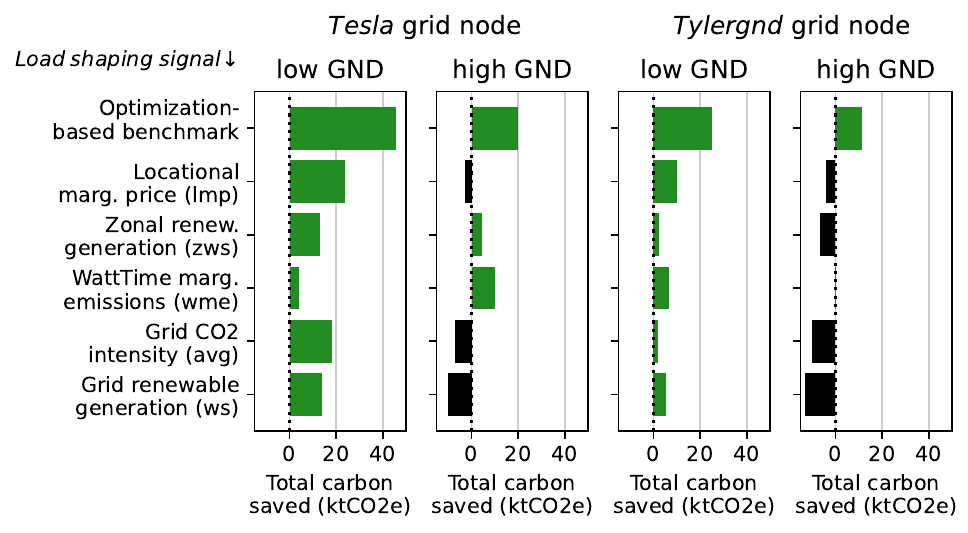}
  \caption{Differences in carbon emissions impact from the shaping approaches in winter and summer months quantified by low and high ERCOT grid net demand (GND) regimes respectively.}
  \label{fig:impactSeasonAlt}
\end{figure}

Using the granular data from the simulation runs, Figure \ref{fig:impactSeasonAlt} shows a clear difference in the effectiveness rankings of load-shaping approaches between months with high and low GND in ERCOT. When the demand is shaped in the Tesla node, the effectiveness of the \textit{lmp} load shaping drops substantially on days with high GND (during summer months in Texas). With higher GND, an approach that ``separates coal from gas'', e.g. \textit{wme}, is better at reducing \ce{CO2} emissions. On the other hand, in the Tylergnd node, it appears that all practicable approaches are worse than keeping load flat. Consequently, we use GND as a feature of interest when examining the efficacy of different load-shaping approaches.

To mathematically formalize these seasonal shifts, daily GND patterns were rigorously classified using the Principal Component Analysis~\citep{abdi2010principal} and then k-means clustering of the obtained daily projections. Interestingly, in case of ERCOT, this method suggested only two clusters, where the high GND regime mainly contains summer days in Texas, spanning June to September. Days in May and October are transition months, and the performance validation discussed later in Subsection~\ref{sec:policy} shows that placing them into either low or high GND clusters does not change the resulting \ce{CO2} emissions savings outcome by more than 2\%.

\subsubsection{Daily $\min(LMP)$}

Using the data from simulations of 365 days in 2023, we identify the features that best predict the load shaping approach with the largest next day reduction of \ce{CO2} emissions. With only 365 daily samples in 2023, using a machine learning algorithm to select daily load shaping risks overfitting, and lacks the physical interpretability required for broad grid application. Therefore, we do not rely on machine learning to act as our final approach selection rule. Instead, we leverage it to help identify which observable grid conditions correlate most strongly with the effectiveness of specific load-shaping approaches. 

Simulation results were used as inputs for training machine learning (ML) models to evaluate the correlation between grid conditions and the effectiveness of load shaping in reducing \ce{CO2} emissions. We extract a multitude of standard, publicly available day-ahead grid signals, including daily minimum, maximum, mean and median LMP, average daily total grid demand, average grid carbon intensity, grid and zonal renewable generation, etc. These features are passed to a Random Forest classifier~\citep{breiman2001random} trained to predict the most carbon-effective shaping approach for each corresponding day. Because Random Forest models utilize bootstrap aggregation and random feature sampling, they are highly effective at isolating feature importance. Upon evaluating the classifier's predictive power, daily minimum LMP ($\min(LMP)$) at the flexible load bus emerged as the dominant predictive feature.

\begin{figure}[h]
  \centering
  \includegraphics[width=0.95\linewidth]{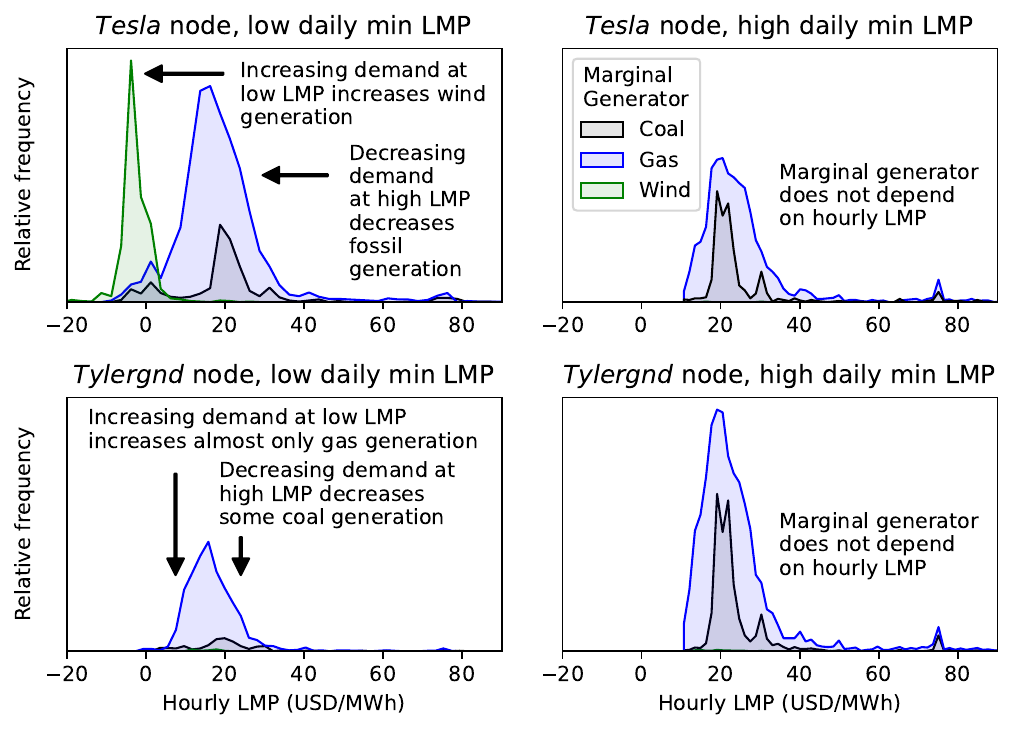}
  \caption{\textit{lmp}-based load shaping is effective only when cleaner technologies are the marginal generator at lower LMPs. In the upper left plot, the Tesla node has a low daily min(LMP), and times of low LMP correspond to wind being a curtailed marginal generator. Following LMP hence increases wind generation and decreases fossil fuel generation. In the lower left plot, following LMP prefers gas over coal dispatch, reducing carbon emissions, although the gains are less significant. The other plots are when daily min(LMP) is higher. Here, hourly LMP barely discriminates coal from gas, so LMP-based load shaping does not impact carbon emissions. In these plots, we define the marginal generator as the technology that meets at least 50\% of the change in total generation (for example, if demand is increased by 80MW, and otherwise curtailed wind increases 60MW, then wind is considered the marginal generator).
  }
  \label{fig:LMPmarginalsplit}
\end{figure}

Having identified $\min(LMP)$ statistically, we sought to establish the physical and economic rationale for its predictive dominance. As visible in Figure~\ref{fig:LMPmarginalsplit} upper left, a very low hourly LMP on the Tesla node (below bid prices of fossil technologies into the day-ahead market, e.g. 5\$/MWh) is a strong predictor of renewable curtailment and renewables being the marginal generation. This is to be expected, since LMP is strongly linked to the bid price of the marginal generator, and renewables often bid 0 or negative prices. Under such conditions, following the LMP signal is effective at reducing \ce{CO2} emissions by increasing demand when curtailed renewables are the marginal generators and decreasing it when fossil fuels are marginal generators. 

In Tylergnd, for the time of this study, gas was economically similar to coal, in the sense that LMP does not definitively select a greater proportion of gas vs a greater proportion of coal generation. Under these conditions, hourly LMP ceases to differentiate generation technologies, and adjusting load in response to \textit{lmp} fails to alter the proportion of gas vs. coal generation and is ineffective at reducing carbon emissions. Since renewables often bid at zero marginal price, a very low LMP is a proxy for renewable curtailment (renewables as the marginal generator). Increasing demand at such hours hence increases renewable generation without any increase in carbon emissions. Thus, it is interesting to examine the combination of grid net demand (GND) in combination with a threshold test of the next day's minimum LMP ($\min(LMP)$) at the flexible load's grid node as a basis for selecting the next day's most effective load shaping approach. 

In view of the above, classifying days into those with high and low $\min(LMP)$ is a useful criterion for selecting a daily load shaping approach. Next, a {\it cherry-picking} strategy is proposed, where the appropriately tuned $\min(LMP)$ thresholds are used to switch between \textit{lmp}, \textit{zws}, \textit{wme} shaping approaches as most effective (demonstrated in Figure \ref{fig:total-impact-over-year}) in reducing \ce{CO2} emissions. In the following sections, we discuss how we ``tune'' the \ce{CO2} emissions-effective {\it cherry-pick} design using the simulation results and analysis. Then, we describe how one can also derive an effective approach selection strategy {\it without requiring a grid simulator}, but using the analysis of the historically observed hour-to-hour market response to grid demand changes.

\subsection{Cherry-picking for effective \ce{CO2} reduction} \label{sec:policy}

Building on the counterfactual simulations and physical insights obtained in Subsection~\ref{sec:features_alt}, we operationalize GND and $\min(LMP)$ into a rule-based dynamic approach selection framework. Rather than applying a single, static load-shaping signal year-round, this framework dynamically selects the most effective approach for the next day based on the anticipated grid operating regime. 

Using the simulation results of the \textit{lmp} and other ``competing'' load shaping approaches (\textit{zws}, \textit{wme}), we \enquote{tune} the $\min(LMP)$ threshold to maximize the impact on the annual savings in grid-level \ce{CO2} emissions. To capture the seasonal differences in grid-level demand - supply matching, we conduct ``threshold tuning'' for low GND and high GND regimes separately. This leads to a four-quadrant cherry-pick rule, i.e. (low GND, low $\min(LMP)$), (low GND, high $\min(LMP)$), (high GND, low $\min(LMP)$) and (high GND, high $\min(LMP)$). By \enquote{sweeping} $\min(LMP)$ threshold values, we calculate the grid-level \ce{CO2} emissions savings by looking at the performance across all approaches in the four quadrants. The most \ce{CO2} effective thresholds are included in Table \ref{tab:thresholds} (NA threshold should be treated as the very large threshold).

\begin{table}[h]
\caption{These were the observed optimal $\min(LMP)$ thresholds in \$/MWh from simulations, from the combinations of GND and $\min(LMP)$.}\label{tab:thresholds}%
\centering
\begin{tabular}{lcc}
\toprule
    & Tesla & Tylergnd \\
\midrule
Threshold A (Low GND) & 2 & NA \\
Threshold B (High GND) & 10 & 18 \\
\bottomrule
\end{tabular}
\end{table}

The optimal rule is validated with leave-one-out validation, a form of cross-validation where one sample is used for validation and the rest is used for model training. Here, this means a cross-validation with 364 days to search for the optimal threshold, and then apply the rule to the 1 day for validation. As a result, we have validation results for every day in the data, which can provide us with a robust view of the \ce{CO2} reductions of the policy.

The best performing \enquote{cherry-picking} strategy derived using the simulation results and the above-described validation process can be summarized as follows:
\begin{algorithm}[h]
\caption{Optimal Cherry-picking Policy}\label{alg:alg1}
\begin{algorithmic}[1]
\State \textbf{if} Low GND (ERCOT winter):
\State \quad \textbf{if} $\min(LMP) <$ LMP Threshold A:
\State \qquad \textit{lmp}
\State \quad \textbf{else}:
\State \qquad \textit{zws} (\textit{lmp} if there are no renewables)
\State \textbf{if} High GND (ERCOT summer):
\State \quad \textbf{if} $\min(LMP) <$ LMP Threshold B:
\State \qquad \textit{lmp} (or \textit{flat}, if there are no renewables)
\State \quad \textbf{else}:
\State \qquad \textit{wme} (or \textit{flat}, if \textit{wme} not available)
\end{algorithmic}
\end{algorithm}

%%%%%%%%%%%%%%%%%%
%%%%%%%%%%%%%%%%%%
\section{Numerical evaluation and discussion} \label{sec:num_eval}

In this section, we evaluate the single-node performance and threshold sensitivity of the cherry-picking strategy (Subsection \ref{sbsec:single_node_eval}), show how the $\min(LMP)$ threshold can be determined directly from observable market data (Subsection \ref{sec:datadriven}), analyze generation dispatch changes during peak net demand hours (Subsection \ref{sbsec:evimpact}), and evaluate two-node spatial and temporal load shifting (Subsection \ref{sec:2nodeshifting}).

\subsection{Single-node cherry-picking performance} \label{sbsec:single_node_eval}

The simple {\it cherry-pick} strategy increases the annual ERCOT's \ce{CO2} emissions savings by a factor of $\sim 2.1$ (with shaping in Tesla) and $\sim 3$ (with shaping in Tylergnd) times when compared to the \textit{lmp} approach alone. With that, the {\it cherry-pick} rule saves $65.6$\% (from shaping at Tesla) and $52$\% (from shaping at Tylergnd) of the upper bound, {\it optimization-based}, benchmark (see Table \ref{tab:PolicyComp}). At the same time, the {\it cherry-pick} strategy makes insignificant change to the average electricity rate in ERCOT: it reduces it by $0.02$\% (with shaping at Tesla), and increases it by $0.9$\% (with shaping at Tylergnd).

\begin{table*}[h]
\caption{\ce{CO2} reduction breakdown showing total strategy impacts}\label{tab:PolicyComp}
\begin{tabular*}{\textwidth}{@{\extracolsep\fill}lcc}
\toprule
Policies & Tesla kt\ce{CO2} reduction & Tylergnd kt\ce{CO2} reduction \\ 
 & \footnotesize{(\% of \textit{optimization-based} benchmark)} & \footnotesize{(\% of \textit{optimization-based} benchmark)} \\
\midrule
\footnotesize{Optimization-based benchmark (upper bound)}  & 65.4 & 36.4 \\
\textit{lmp} & 20.8 (32\%) & 6.3 (17\%) \\ 
\textit{wme} & 10.2 (16\%) & 1.6 (4\%) \\
\textit{zws} & 17.8 (27\%) & -4.3 (-12\%) \\
\textit{avg} & 10.9 (17\%) & -8.8 (-24\%) \\
\textit{cherry-pick} & 42.9 (66\%) & 18.9 (52\%)\\
\bottomrule
\end{tabular*}
\vspace{1ex}
\footnotesize{Note: \ce{CO2} reduction is computed with respect to the grid-level \ce{CO2} emissions when Tesla/Tylergnd load is ``flat'' (i.e. at baseline).}
\end{table*}

We evaluate the sensitivity of the total \ce{CO2} emissions savings to changes in $\min(LMP)$ thresholds. While the maximum \ce{CO2} emissions saving is achieved at ($\min(LMP)$ @ low GND threshold, $\min(LMP)$ @ high GND threshold) = ($2$, $10$), this impact is not very sensitive to changes in the threshold. For example, the differences in \ce{CO2} emissions savings across thresholds $\min(LMP)$ @ low GND threshold within interval $[-3, 5]$ and $\min(LMP)$ @ high GND threshold within interval $[8, 13]$ do not differ from each other more than $12$\%. This holds for both Tesla and Tylergnd bus locations.

Note that the forecasting accuracy of DA $\min(LMP)$ and other day-ahead grid signals (such as zonal wind and solar generation) can affect the choice of load shaping approach and hourly schedule for the next day. While here we do not explicitly simulate forecast errors, the low sensitivity of the total \ce{CO2} emissions savings to the selected $\min(LMP)$ threshold provides a direct robustness guarantee against day-ahead LMP forecasting uncertainty: because annual \ce{CO2} reductions vary by less than $12$\% across a wide band of threshold values ($[-3, 5]$\,\$/MWh in low GND and $[8, 13]$\,\$/MWh in high GND), moderate forecast errors near the threshold boundary only occur on days when the competing approaches achieve comparable emissions impact. Moreover, because the max-nominal-min shaping policy (Figure \ref{fig:policy}) depends only on the ordinal ranking of hourly signal values within the day (identifying the $9$ highest- and $9$ lowest-signal hours) rather than exact point magnitudes, the resulting daily load schedules are inherently resilient to uniform bias or moderate level errors in day-ahead price and renewable generation forecasts.

Next, we discuss how one can gain insights on marginal generation by looking at the historically cost-effective outcomes of the day-ahead market dispatch, and using them to derive the emissions-effective {\it cherry-pick} strategy. 

\subsection{Determining the $\min(LMP)$ parameter from observable grid data} \label{sec:datadriven}

To this point, the cherry-picking rule and parameters have been verified using statistically
derived ERCOT GND (refer to Section \ref{sbsec:gnd}), but with a simulation-determined $\min(LMP)$ threshold. To scale applicability to other grids and to common scenarios where the grid simulator is not available, the $\min(LMP)$ threshold parameter needs to be determined from the commonly available data. This requires understanding the impact of nodal demand change given nodal $\min(LMP)$.

In general, it is difficult to identify the impact of the nodal demand change using the available grid dispatch data. The main reason for this is the complexity of the electricity grid as a system, where the OPF outcomes require understanding grid infrastructure and grid operations in detail. We show, however, that it is possible to infer $\min(LMP)$ via (i) relating the local price at the controllable demand node to GND, and (ii) drawing conclusions from the historical, cost-effective, day-ahead dispatch response to hour-to-hour increases in grid demand.

First, data observations show that the nodal $\min(LMP)$ price of both Tesla and Tylergnd nodes has a {\it linear trend} with GND, i.e. on low GND days (winter in ERCOT) their $\min(LMP)$ tend to be lower as well, and vice versa. The high/low GND regimes separate ERCOT's economic supply-demand matching (via DC-OPF) into operating scenarios of high and low grid net demand, not matched by the cheap intermittent renewable generation. These regimes are different in the way that the ERCOT operator ends up accommodating changes in demand depending on its location, which is significantly affected by the proximity and generation capacity of renewable and fossil fuel generators, their dynamics and the underlying transmission infrastructure.

Next, we observe how frequently DC-OPF uses a particular generation technology (i.e. wind \& solar, gas, coal) to supply an hour-to-hour grid demand increase at Tesla bus. Using this information, local (zonal) generation and congestion, and renewable curtailment, we outline how to cherry-pick among the load shaping approaches in order to reduce grid-level \ce{CO2} emissions. Analogous reasoning can be used to construct the {\it cherry-pick} strategy for shaping load at Tylergnd bus.

\begin{figure}[h]
  \centering
  \includegraphics[width=\linewidth]{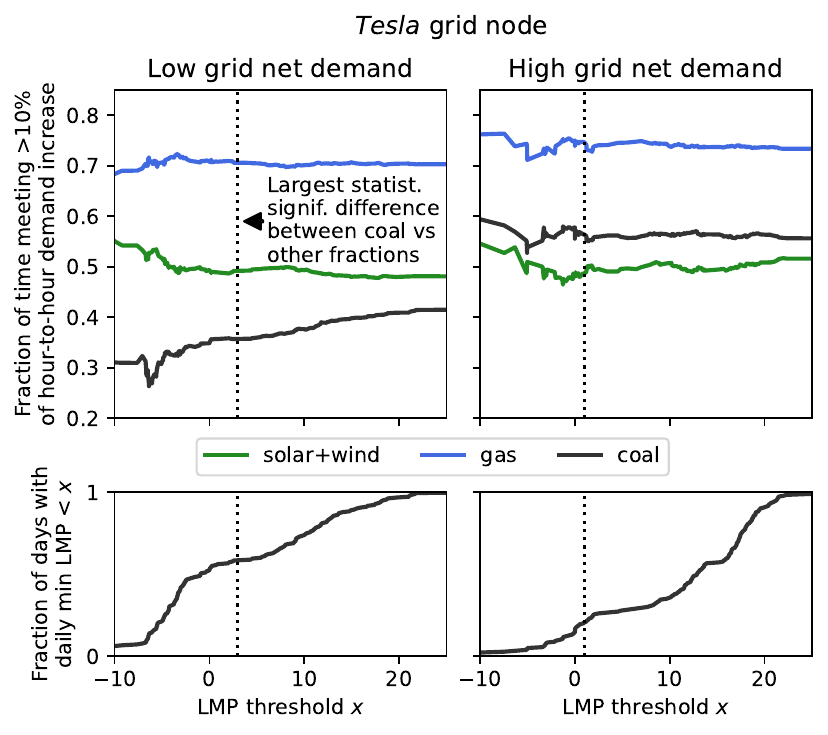}
  \caption{Dominant generation changes at times of increased demand in Tesla bus. Dotted lines match $\min(LMP)$ threshold with the largest statistically significant ratio between non-coal (wind, solar and gas) and coal supply contributions to demand growth in Texas.}
  \label{fig:TeslaIncrease}
\end{figure}

\begin{figure}[h]
  \centering
  \includegraphics[width=\linewidth, trim = 0cm 8cm 7cm 0cm, clip]{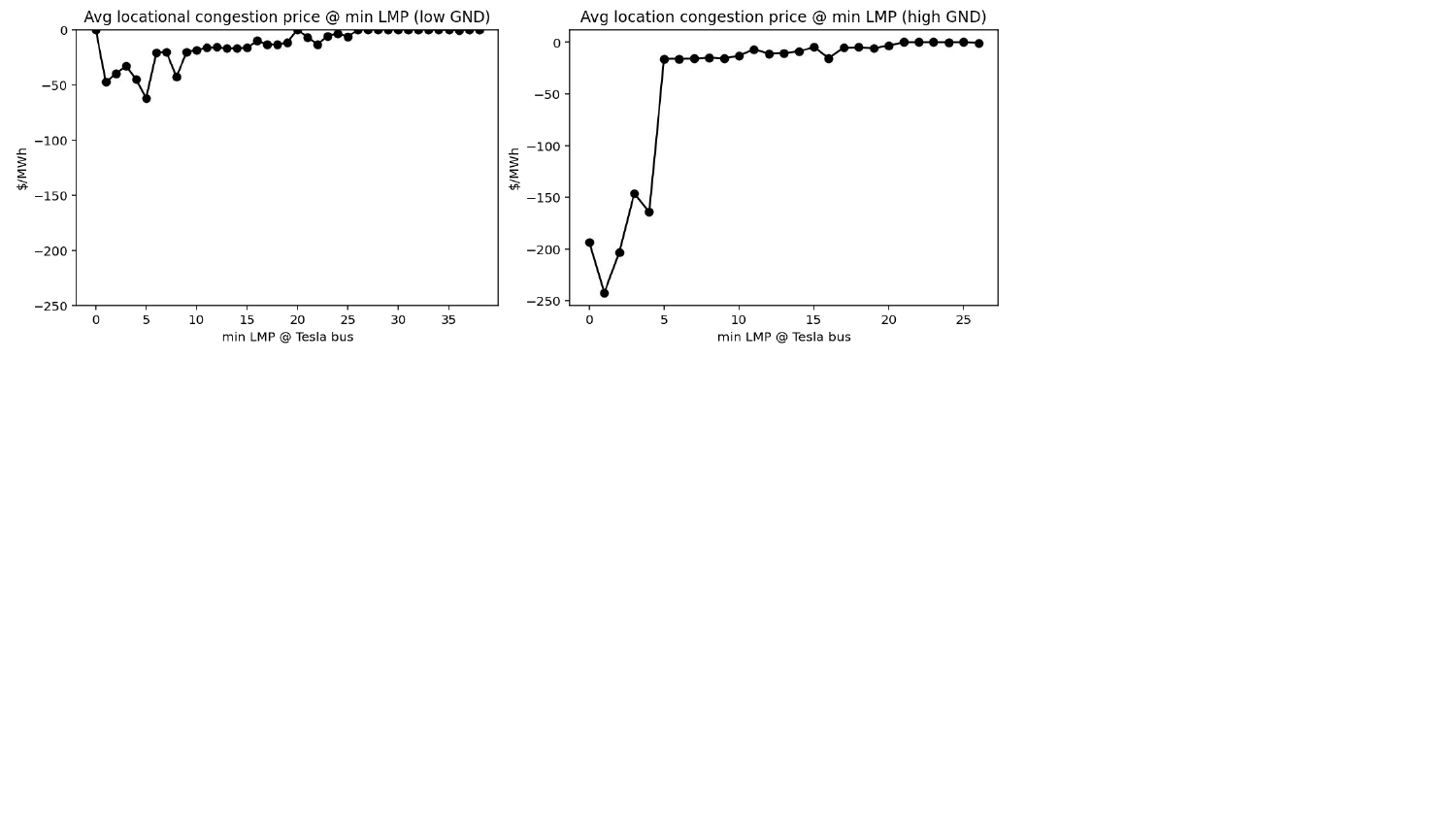}
  \caption{{Marginal congestion price at Tesla bus as a function of its daily $\min(LMP)$.}}
  \label{fig:TeslaMCC}
\end{figure}

\begin{figure}[h]
  \centering
  \includegraphics[width=\linewidth, trim = 0cm 0cm 0cm 0cm, clip]{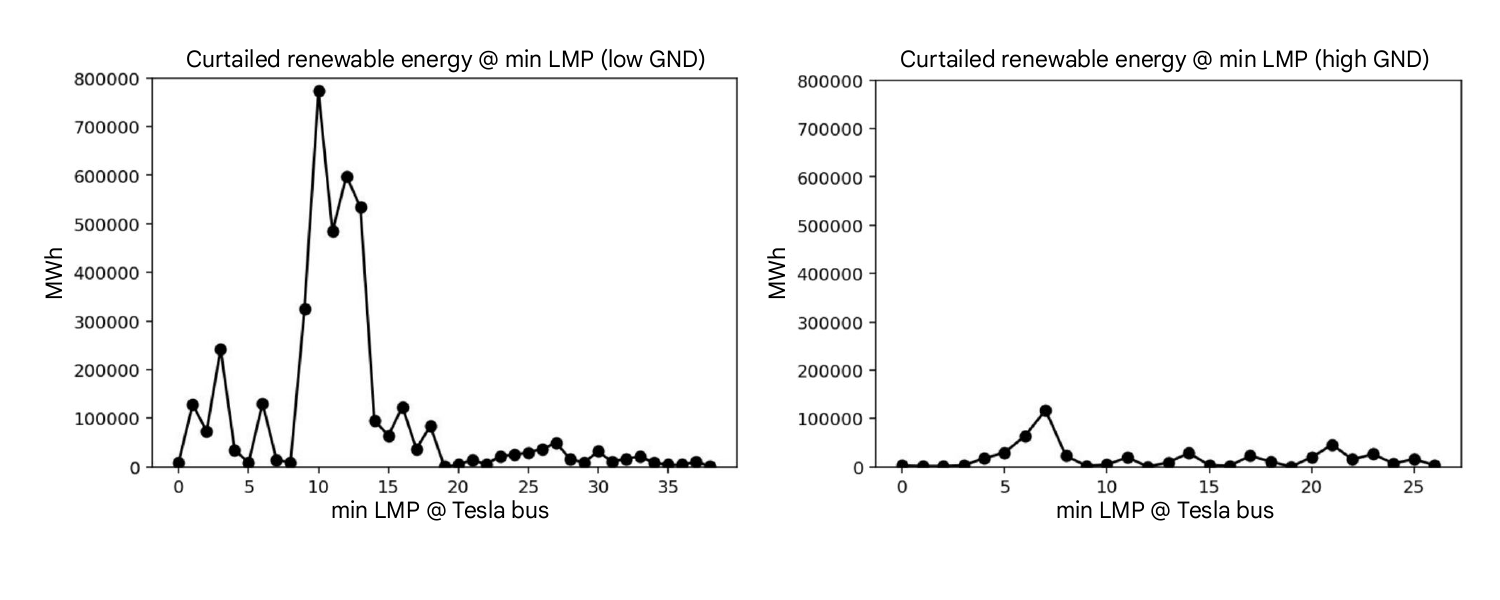}
  \caption{Curtailed renewable energy as a function of $\min(LMP)$ at Tesla bus.}
  \label{fig:TeslaCurtailment}
\end{figure}

On low GND and low $\min(LMP)$ days, it is cost-effective (i.e. driven by the economic dispatch) to supply demand growth from renewable and gas generation (see left side in Figure \ref{fig:TeslaIncrease}), which means that following \textit{lmp} approach is emissions-effective as well. With higher daily $\min(LMP)$ during low GND season, coal is used more often to supply increases in grid demand. Based on negative marginal congestion prices at Tesla bus (Figure \ref{fig:TeslaMCC}) and notable curtailed renewable energy (Figure \ref{fig:TeslaCurtailment}), we conclude that following abundant renewables (using \textit{zws} approach) increases the chance of their dispatch when the Tesla demand is being controlled. 


On high GND days, DC-OPF's economically-preferred generation contributions to the grid demand growth changes. The dominating technologies in this case are gas and coal (right in Figure \ref{fig:TeslaIncrease}). Negative marginal congestion prices and the existing grid curtailment when $\min(LMP)$ is low (Figures \ref{fig:TeslaMCC} and \ref{fig:TeslaCurtailment}) suggest that using \textit{lmp} approach could reduce the existing curtailments. Note, however, that $\min(LMP)$ is high for the majority of days during summer months in Texas, and that gas and coal are the dominant technologies to supply hour-to-hour growth in grid demand. On these days, load shaping approaches that are able to separate hours with predominantly gas vs coal activity at the margin (e.g. \textit{wme}) are more effective in reducing the dispatch emissions as well.

Based on Figure \ref{fig:TeslaIncrease}, the largest statistically significant ratio between non-coal (wind, solar and gas) and coal supply contributions to demand growth occurs on days with $\min(LMP)$ smaller than $3$\,\$/MWh (low GND, in Texas winter) and $1$\,\$/MWh (high GND, in Texas summer). These thresholds are below bid prices for fossil technologies (which are rarely lower than $10$\,\$/MWh), suggesting that (at least for some hours in the day), the marginal generator is one with negligible marginal cost, primarily renewables. Indeed, these days have a high rate of wind curtailments, indicating that following the \textit{lmp} approach (increasing demand at times of low LMP) may increase dispatch from wind generators and reduce their energy curtailment. Thus, we propose the following cherry-pick strategy for choosing between load shaping approaches in Table \ref{tab:TeslaRule}.

\begin{table*}[h]
\caption{Strategy and carbon savings from the threshold-based load shaping approach selection @ Tesla grid node}\label{tab:TeslaRule}
\begin{tabular*}{\textwidth}{@{\extracolsep\fill}lccc}
\toprule
 & winter & summer & Total kt\ce{CO2} saved \\ 
 & \footnotesize{Threshold $= 3$} & \footnotesize{Threshold $= 1$} & \footnotesize{(\% of the simulation benchmark)} \\
\midrule
$\min (\textit{LMP}) <$ Threshold & \textit{lmp} & \textit{lmp} & \multirow{4}{*}{41.45 (63\%)} \\
 & 24.5 & 5.95 & \\
\cmidrule{1-3}
$\min (\textit{LMP}) \geq$ Threshold & \textit{zws} & \textit{wme} & \\ 
 & 2.2 & 8.8 & \\
\bottomrule
\end{tabular*}
\vspace{1ex}
\footnotesize{Note: Thresholds are based on transitions in the grid supply regime (based on Figure \ref{fig:TeslaIncrease}) and based on $\min (\textit{LMP})$ values above which grid curtailments are close to zero. Note that the annual \ce{CO2} savings are only $4$\% smaller than the tuned simulation performance in Table \ref{tab:PolicyComp}.}
\end{table*}

\subsection{The impact of price-driven load shaping approaches on generation dispatch} \label{sbsec:evimpact}

Because the price-driven \textit{lmp} approach serves as a primary building block of the cherry-picking policy evaluated in Subsections~\ref{sbsec:single_node_eval}--\ref{sec:datadriven} and is widely adopted by price-responsive distributed energy resources, we next examine its physical generator-dispatch mechanism during peak net demand hours and contrast it with naive time-of-use (\textit{overnight}) price response. The temporal flexibility of electric vehicle (EV) charging has been the focus of many studies, where charging stations and EV fleets managed by flexible load aggregators, or operated as Virtual Power Plants (VPPs), shape their demand to reduce energy cost (see, e.g., \citep{bessa2012economic, bayram2014pricing, fan2020optimal}). As an example, here we model and simulate the cost-effective \textit{overnight} charging approach where we shed load during high demand hours during a day (noon - 9pm), and increase it during nighttime (10pm - 7a.m. of next day). The simulation results for this type of shaping of Tesla's and Tylergnd's demand show that the \textit{overnight} approach is almost as effective as the \textit{lmp} approach in terms of electricity cost savings, but with increased grid-level \ce{CO2} emissions (see Figure \ref{fig:total-impact-over-year}) due to an increase in dispatch of cheap coal generation.

Note that the demand shaping that follows the \textit{lmp} shaping resembles load shedding at peak (net) demand events, since times of high LMPs are typically those with high grid net demand as well. We evaluate the {\it average hourly change} in dispatched generation between \textit{lmp} and \textit{baseline} (\enquote{flat} load) during the $100$ hours with the highest GND. With changes in Tesla bus, there is primarily a drop in non-combined cycle gas (46MW), coal (22MW) and battery dispatch (24MW) with a slight increase of combined cycle gas. In case of Tylergnd, we observe a drop in combined cycle gas (26MW), coal (18MW), non-combined cycle gas (20MW) and batteries (20MW). Using the simulation results, we also compute that the \textit{lmp} approach reduces peak GND values by around 80MW (consistent with an 80MW load drop). It also indicates that load flexibility may reduce the required grid generation and storage capacity to meet the same energy demand, which is consistent with other studies \citep{norris2025rethinking, Au2025DemandResponseSPP}.

%%%%%%%%%%%%
%%%%%%%%%%%%
\subsection{Temporal and spatial shifting of energy demand between data centers} \label{sec:2nodeshifting}

\begin{figure}[h]
  \centering
  \includegraphics[width=\linewidth]{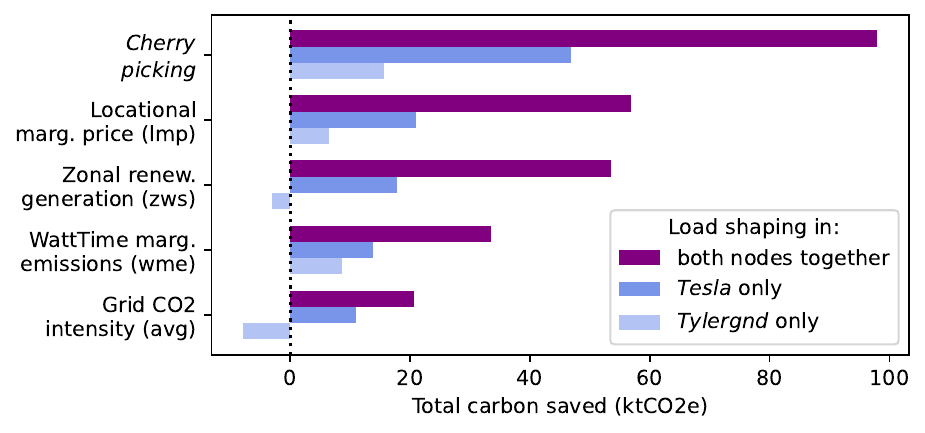}
  \caption{\ce{CO2} savings from temporal and spatial shifting between Tesla and Tylergnd buses.}
  \label{fig:twopolicy}
\end{figure}

Large compute providers, such as Google, Meta, Amazon, Microsoft Azure, etc., have data centers scattered across different grids with varying carbon intensity. Workload location and start time is generally managed via proprietary schedulers that leverage the distributed computing, storage, accelerator platforms and communication network. This network can transfer workloads, and hence energy consumed, across time and also between locations \citep{kelly2016balancing}. This special capability can further improve the impact of multi-data center load shaping on grid-level \ce{CO2} emissions.

We evaluate the impact of spatial and temporal shifting by modeling two large data center loads, one located at Tesla and the other at Tylergnd bus. Constraints are similar to the case of a single data center demand shaping: the baseline regime assumes ``flat'' power demand of 400MW for both data centers; no data center can draw more than 480MW or less than 320MW power. Since workload can be shifted from one data center to another, the total energy demand constraint covers both, totalling $2\times24h\times400$MW. We simulate 2-node \textit{lmp}, \textit{zws}, \textit{wme}, \textit{avg} load shaping using the analogous approach as in the single bus case, i.e. we increase or decrease hourly power demand at $2$ nodes based on the ranking of the signal (e.g. LMP) across the 48 daily hours of the 2 locations. For \textit{lmp}, \textit{wme} and \textit{avg}, we decrease demand at the highest signal values, and do the opposite for \textit{zws}. Note that, in the case where the load shaping signal is the same for both Tesla and Tylergnd (for \textit{wme} and \textit{avg}), the outcome of the greedy energy allocation will be the same as if both locations were shaped independently, i.e. there will not be spatial shifting between the locations. For approaches following signals that differ for Tesla and Tylergnd, such as \textit{lmp} and \textit{zws}, the greedy energy allocation will prefer locations and hours with the lowest LMP and the largest zonal renewable energy production. 

The conducted two-node simulations replayed over the whole year 2023 reveal significantly improved \ce{CO2} emissions savings across all approaches (see Figure \ref{fig:twopolicy}). Spatial shifting of energy consumption to ``greener'' and ``cheaper'' grid zones more effectively reduces grid-level \ce{CO2} emissions in that they are able to leverage higher renewable energy production of the ``greener'' zones and shift some load away from mainly fossil fuel supplied zones. Using the higher effectiveness of Tesla demand shaping (see Table \ref{tab:PolicyComp}), we also evaluate the two node {\it cherry-pick} strategy, where switching between two-node load shaping approaches is driven by Tesla's optimized strategy and thresholds in Table \ref{tab:thresholds}. The simulation results show that this two-node {\it cherry-pick} strategy reduces more than 70\% more grid-level \ce{CO2} emissions when compared to the two-node \textit{lmp} approach.

The main rationale for switching between load shaping signals in case of a multi-node spatial and temporal energy shifting is based on the same trade offs as discussed in the previous sections. Shifting energy to a zone with plentiful, curtailed and cheap renewable generation from a zone rich in fossil fuels would decrease both the electricity cost (total cost of cleared energy in the market) and renewable curtailments and, thus, reduce \ce{CO2} emissions from the dispatched generation. Only in the high demand regimes, with no excess of renewable capacity, where the dispatcher seeks to find the cost effective interplay between coal and gas generation, the marginal signals that most effectively separate between coal and gas also reduce the grid-level emissions the most and shifting demand spatially might not result in larger \ce{CO2} emissions reductions.

\section{Conclusions and future work} \label{sec:conclusions}

The goal of this research is to derive a simple, implementable and explainable day-ahead load shaping strategy that effectively reduces grid-level \ce{CO2} emissions with minimal impact on cost of electricity (i.e. total energy cost in the day-ahead market). For our investigations involving ERCOT's Tesla and Tylergnd buses, it is found that a rule-based {\it cherry-picking} strategy, where a daily choice is made from a subset of commonly considered load shifting approaches, outperforms any individual one by 2 to 3 times in terms of reduced grid-level \ce{CO2} emissions. The impact of load shaping on dispatched generation is based on counterfactuals, which is then analyzed with regards to principal grid, zonal and nodal characteristics. By exploring the effectiveness of different load shaping approaches as a function of grid operating conditions, and then recognizing those conditions, a simple strategy is derived for cherry-picking the next day's load shaping approach. In our simulations, we consider a flexible load with parameters (minimum, base and peak power) based on a data center. However, results also apply to any other load (such as a flexible factory or VPP) with similar parameters.

The emissions impact of daily approach cherry-picking was found to be dependent on flexible load's grid location, which should be carefully considered when selecting a location for data center, storage or, generally, VPP, that are able to modulate their power consumption in the considered fashion. Collocation or proximity to curtailable intermittent generation is preferable, since remote power generation is subject to transmission congestion and also price elevation from transmission. Proximity to cheap coal, on the other hand, can result in net increases in \ce{CO2} emissions. Motivated by compute shifting between data centers, we simulate the load shaping approaches that prioritize locations and times with the lowest locational marginal price, highest zonal wind and solar, or lowest zonal WattTime \citep{watttime2021watttime} signal when allocating computing and its associated energy. The simulation results reveal that spatial and temporal shifting of energy consumption to ``greener'' and ``cheaper'' grid zones and times can reduce grid-level \ce{CO2} emissions more than 2 times when compared to the temporal shifting only, with boosted \ce{CO2} emissions reductions when cherry-picking is used to sample from two-node shifting approaches. 

To the best of our knowledge, this paper is the first to use counterfactual simulation to provide insights into the effectiveness of different typically discussed load shaping approaches on dispatched electricity generation, \ce{CO2} emissions and cost of electricity, and uses the derived insights to design the strategy for cherry-picking next day's load shaping approach. From a policy and carbon-accounting perspective, our results also highlight a critical misalignment between administrative carbon footprints and physical grid decarbonization. Specifically, shaping load using grid-average carbon intensity (\textit{avg})---the standard signal under Location-Based GHG Protocol Scope 2 accounting \citep{GHGProtocol2015Scope2}---captures only 25\% of the \ce{CO2} reductions achieved by the proposed \textit{cherry-pick} strategy at the renewable-rich Tesla node, and actively backfires at the fossil-dominated Tylergnd node by increasing system-wide \ce{CO2} emissions by 8.8\,kt\ce{CO2}/yr relative to a flat load profile. This demonstrates that corporate sustainability frameworks and Scope 2 accounting standards that evaluate flexible loads solely against region-wide average emission rates can incentivize operational behaviors that increase physical grid emissions, underscoring the need to incorporate locational, price-aware, and congestion-aware signals into sustainability guidelines.

Areas for future research investigation include: (i) analogous analyses in other grids with different energy dispatch operations; (ii) analysis for avoidance of conditions under which nuclear power generation is negatively impacted by load shaping, in which its slow ramp rate can lead to an increase in system \ce{CO2} emissions and increase in cost from other generation sources; (iii) studying effects of different levels of flexibility and the result of various infrastructure upgrades and changes in grid generation portfolio; and (iv) evaluating fleet-scale market feedback when multiple gigawatts of data centers or VPPs shift load simultaneously. While a single 400\,MW flexible load induces negligible wholesale market price changes ($\le 0.6$\,\$/MWh), uncoordinated multi-gigawatt load shifting in response to the same guiding signal could exhaust local renewable curtailment, alter transmission congestion patterns, and shift the identity of the marginal generator, thereby creating endogenous feedback on nodal LMPs and marginal emission rates. Extending the proposed cherry-picking framework to account for aggregate fleet-level market feedback and distributed coordination across large populations of flexible loads remains an important direction for future work. The expectation is that the fundamental approach and validations derived in this paper will apply, and that the collective, distributed carbon-aware load shaping using the derived unified strategy is scalable and effective for \ce{CO2} emissions reduction. Note, however, that the design of the \ce{CO2} emissions-reducing load shaping directly ties to the cost structure across different generation technologies and the associated market clearing process. And, as demonstrated in this study, strategy parameters where grid behavior is in alignment with \ce{CO2} reduction can be inferred using the historical dispatch results. When this is not the case, then the outcome could be an increase in \ce{CO2} emissions and market price.

\section*{Acknowledgements}
The authors would like to thank Olivier Corradi and Julien Lavalley from Electricity Maps, and Neema Zadeh, John Platt and Line Roald at Google for their invaluable feedback and help in improving the quality of the paper. We would also like to thank the Advanced Energy team at Google for funding this research.

\appendix

\section{Daily causal simulation approach} \label{sec-appendix}
{\it MarketnSight}, developed by Electric Power Engineers (EPE) \citep{EPEConsulting}, is a Security Constrained Economic Dispatch (SCED)-based market analysis simulator, calibrated to accurately emulate ERCOT's day-ahead DC Optimal Power Flow (DC-OPF) economic dispatch process. MarketnSight iteratively solves a large-scale, fundamentally nonlinear, mixed-integer Security Constrained Optimal Power Flow optimization, i.e.

\begin{equation*}
\begin{aligned}
\min \quad & f_0(x_0,u_0) && \text{(*)} \\
\text{subject to} \quad & g_0(x_0,u_0) = 0 && \text{(1)} \\
& h_0(x_0,u_0) \le L_l && \text{(2)} \\
& g_k^s(x_k^s,u_0) = 0, \quad k = 1, \dots, c && \text{(3)} \\
& h_k^s(x_k^s,u_0) \le L_s, \quad k = 1, \dots, c && \text{(4)} \\
& g_k(x_k,u_k) = 0, \quad k = 1, \dots, c && \text{(5)} \\
& h_k(x_k,u_k) \le L_m, \quad k = 1, \dots, c && \text{(6)} \\
& |u_k - u_0| \le \Delta u_k, \quad k = 1, \dots, c && \text{(7)}
\end{aligned}
\end{equation*}

where the objective function $f_0(x_0,u_0)$ represents the base-case generation dispatch costs, with $x_0$ being the state variables (e.g. voltage magnitude, {\it load}) and $u_0$ control variables (e.g. generators' active power, terminal voltage, controllable transformer ratios, and breaker statuses). Pre-contingency constraints, (1) and (2), represent DC power flow equality constraints and inequality constraints for normal operating limits, such as physical equipment limits. Post-contingency constraints, (3) to (6), ensure that the system remains within short and medium-term emergency limits after a contingency. Coupling constraints (7) set the maximum allowed adjustment of control variables between the base case and the $k$-th post-contingency state. This is determined by the available time for corrective actions and the rate of change of the equipment. To evaluate the impact of different load shaping approaches on generation dispatch, we replace flexible load values (subset of $x_0$) at targeted bus nodes with the precomputed values based on the particular load shaping approach.

Note that the contingency is a postulated event or scenario that disrupts the normal operation of the electric grid, e.g. transmission line, transformer, or generator failure, or unexpected variations in power injections, such as a sudden drop in wind or solar generation, or data center power demand.

MarketnSight solves strictly DC SCED for day-ahead dispatch used in ERCOT, where the problem focuses purely on active power re-dispatch and line flows, which is much faster to solve using Linear Programming or Mixed Integer Linear Programming (MILP). Day-ahead optimization incorporates a Unit Commitment, which expands the problem into a Security Constrained Unit Commitment (SCUC). An SCUC solver minimizes total bid costs over the entire time horizon while simultaneously enforcing the network base case and contingency constraints for every hour. The standard SCED (*) assumes a fixed schedule of which generators are turned on or off. Incorporating UC means the optimization model actively decides the start-up, connection, and disconnection schedule of generating units across the day. Because generator statuses are \enquote{on/off} decisions, the optimization problem is a large-scale Mixed Integer Non-Linear Programming (MINLP) problem (or an MILP if a DC approximation is used) \citep{luo2010security}. 

\subsection{Optimization-based benchmark}
\label{sbsec:simulation}
We use the granular DC-OPF simulator of ERCOT's day-ahead generation dispatch process to compute the optimal \ce{CO2} emissions-aware flexible load shapes and use these as the benchmark for comparing the impact across the considered load shaping approaches. To achieve this, we add to the DC-OPF optimization objective (*) total \ce{CO2} emissions of the dispatched generation multiplied by the penalty term $\lambda$ which we treat as a parameter and change it to explore the impact of different tradeoffs between the economic and emissions goals. We also treat the flexible load demand in the selected buses as control variables, where we add the constraints which enforce the total load daily energy to be equal to \(400\times 24 = \SI{9600}{\mega\watt h}\), and that the load must consume between \SI{320}{\mega\watt} and \SI{480}{\mega\watt}. Then, for each day in 2023, the flexible 24-hour load shape that co-optimizes the dispatch cost and the penalty-weighted dispatched \ce{CO2} emissions was computed for different penalties $\lambda$. Among the computed optimal load shapes for different $\lambda$s, the optimization-based benchmark is selected to be the shape that achieves the minimum daily \ce{CO2} emissions when used with purely economic DC-OPF dispatch, i.e. the day-ahead market settlement process without the altered objective and constraints \citep{luo2010security}.

\bibliographystyle{elsarticle-num} 
\bibliography{sn-bibliography}

\end{document}